\documentclass{article}

\usepackage{microtype}
\usepackage{graphicx}
\usepackage{subfigure}
\usepackage{booktabs} 
\usepackage{xspace}

\usepackage{hyperref}

\usepackage{verbatim}
\usepackage{tabularx}
\usepackage{algorithm2e}

\newlength\mylen

\usepackage{amsmath}

\usepackage{cprotect}

\usepackage{listings}
\usepackage{color, colortbl}
\definecolor{codegreen}{rgb}{0,0.6,0}
\definecolor{codegray}{rgb}{0.5,0.5,0.5}
\definecolor{codepurple}{rgb}{0.58,0,0.82}
\definecolor{backcolour}{rgb}{0.95,0.95,0.92}
\lstdefinestyle{mystyle}{
   backgroundcolor=\color{backcolour},
   language=C++,
   commentstyle=\color{codegray},
   keywordstyle=\color{blue},
   numberstyle=\small\color{codegray},
   stringstyle=\color{codepurple},
   basicstyle=\ttfamily,
   breakatwhitespace=false,        
   breaklines=true,                
   captionpos=b,                    
   keepspaces=true,                
   numbers=left,                    
   numbersep=6pt,                  
   showspaces=false,                
   showstringspaces=false,
   showtabs=false,                  
   tabsize=2
}
\usepackage[T1]{fontenc}
\usepackage{lmodern}
\newcommand{\recompile}{\textsc{recompile}\xspace}
\newcommand{\recompileNoS}{\textsc{recompile}\xspace}
\newcommand{\dynamoe}{\textsc{DynaMoE}\xspace}
\newcommand{\dynamoeNoS}{\textsc{DynaMoE}\xspace}

\newcommand{\mycomment}[1]{}

\usepackage[accepted]{mlsys2022}

\mlsystitlerunning{DynaMoE: Dynamically optimized Mixture of Experts}

\begin{document}

\twocolumn[
\mlsystitle{Optimizing Mixture of Experts using Dynamic Recompilations}



\mlsyssetsymbol{equal}{*}

\begin{mlsysauthorlist}
\mlsysauthor{Ferdinand Kossmann}{eth}
\mlsysauthor{Zhihao Jia}{cmu}
\mlsysauthor{Alex Aiken}{stanford}
\end{mlsysauthorlist}

\mlsysaffiliation{eth}{ETH Zurich}
\mlsysaffiliation{cmu}{Carnegie Mellon University}
\mlsysaffiliation{stanford}{Stanford}

\mlsyscorrespondingauthor{Ferdinand Kossmann}{ferdiko@mit.edu}

\mlsyskeywords{Mixture of Experts, MoE, recompile, dynamic optimization, FlexFlow, sparse machine learning}

\vskip 0.3in

\begin{abstract}
The Mixture of Experts architecture allows for outrageously large neural networks by scaling model parameter size independently from computational demand (FLOPs). However, current DNN frameworks cannot effectively support the dynamic data flow in Mixture of Experts, and implementations on top of these frameworks need to use workarounds that introduce significant overheads. To address the limitation of these frameworks, we present \dynamoe, a DNN library that uses {\em dynamic recompilations} to optimize and adapt the use of computational resources to the dynamic needs of Mixture of Experts models. Our evaluation shows that \dynamoe achieves a $1.8\times$ speedup and supports $2.3\times$ larger model sizes when compared to existing MoE systems, even when not using recompilations. We then present further optimizations enabled by dynamic recompilations that yield an additional $1.7\times$ speedup while simultaneously reducing memory pressure and improving model quality.
\end{abstract}
]





\printAffiliationsAndNotice{}

\section{Introduction}
\label{sec:introduction}

The scaling of model parameters and training sample size is one of the most effective paths towards more powerful models \cite{scaling, fewshot}: simple neural architectures with many trainable parameters and large amounts of training data generally outperform more complicated neural networks that are trained at a smaller scale~\cite{bitterlesson}. However, both the increase of model parameters and sample size introduce additional computation, imposing a bottleneck to further scaling.

Mixture of Experts (MoE) addresses this computational bottleneck by allowing model parameters and computational demand (FLOPs) to be scaled independently. This allows to further scale the model parameter size by several orders of magnitude: MoE models like Switch transformer demonstrate that MoEs make scaling beyond billion model parameters feasible --- Switch transformer hereby has 5000$\times$ more parameters than BERT. ST-MoE is currently leading most NLP benchmarks \cite{st-moe} and V-MoE is matching the state-of-the-art predictive performance on image tasks with as little as half of the compute at inference time \cite{vmoe}.

In order to support such large parameter sizes, MoEs only activate submodules of the network for each sample, so-called {\em{experts}}. Which experts to use for a given sample is learned by a so-called \textit{gating network}. This introduces dynamic data flow, where the inputs to an operator are only determined at runtime and change over the course of training.

Current machine learning systems however still cater to static workloads and struggle with the dynamic data flow in MoEs.  For example, frameworks such as TensorFlow 
require a user to define computation graphs with static tensor sizes that cannot be changed during training \cite{tensorflow}. 
To deal with variations in the amount of samples that are assigned to an expert, users need to employ workarounds that lead to significant runtime overheads and suboptimal model quality.

%

In this paper, we introduce \recompile, a simple mechanism that enables dynamic changes to a model architecture and its tensor sizes during training. \recompile enables two key systems optimizations that current MoE frameworks lack. First, by using \recompile, we introduce {\em dynamic capacity factors} that sporadically adapt the tensor sizes of each expert based on their actual need, reducing the computational demand and memory requirement of the MoE. The second optimization leverages that the sample assignments of the gating network converge early in training. Based on this observation, we introduce {\em sample assignment caching}, which caches the sample assignments after their convergence and uses \recompile to eliminate the dependency between the gating network and the experts, resulting in more parallelism.

To evaluate \recompile and our MoE optimizations, we implement \dynamoe, an open-source MoE training system with a flexible API and rich support for popular operators.
 
In summary, our contributions are:

\begin{itemize}
    \item We present \dynamoeNoS, a publicly available implementation of MoE. Compared to state-of-the-art public MoE systems, \dynamoe delivers speed ups of up to $1.8\times$ while supporting $2.3\times$ larger models before running out of memory.
    \item We propose \recompileNoS, a mechanism that can be added to any ML framework to handle the dynamic data flow in models such as MoEs. We showcase how \recompile naturally enables two optimizations for MoEs that yield an additional speedup of $1.7\times$. These optimizations not only reduce runtime but also reduce memory pressure and improve model quality. 
\end{itemize}


\section{Background \& Related Work}
\label{sec:rel_work}

We briefly discuss previous work on both the algorithmic ideas underlying MoEs as well as systems that aim to scale MoEs. 

\subsection{Algorithmic work}
Mixture of Experts was originally proposed in \cite{origpaper}. The key idea behind MoEs is to build a network consisting of multiple submodules, so-called {\em experts}. Each expert should hereby specialize on a subdomain of the input, assuming that solving only that subproblem is easier than solving the whole problem altogether. To decide which experts are the most competent for a given sample, a {\em gating network} assigns weights to each expert's prediction, and the final MoE prediction is given by the weighted sum of the expert predictions. 

\citet{origpaper} introduce multiple loss functions for training an MoE, among which we present two: \textit{cooperation loss} and \textit{specification loss}.
The cooperation loss $\mathcal{L}_{coop}$ is given by Equation~\ref{eq:cooploss}.

\begin{equation}
\mathcal{L}_{coop}(\hat{y}, O, g)=\mathcal{L}(\hat{y}, \sum_{i} g_{i} O_{i})
\label{eq:cooploss}
\end{equation}

where $\mathcal{L}$ is any task-specific loss function (e.g. categorical cross-entropy), $O_i$ is the sample prediction of the $i$-th expert, $O=(O_1, ..., O_n)$ are the sample predictions of all experts, $g_i$ represents the weight assigned to the $i$-th expert, and $\hat{y}$ is the sample's label.

The cooperation loss is therefore just the task-specific loss between the target vector $\hat{y}$ and the MoE output $\sum_{i} g_{i} O_{i}$. Note that the cooperation loss incentivizes experts to cooperatively produce a good prediction: The gradients to each expert's network in back propagation also depend on the predictions made by the other experts during the forward pass.

Alternatively, the specification loss is given by Equation \ref{eq:specloss}.

\begin{equation}
    \mathcal{L}_{spec}(\hat{y}, O, g)=\sum_{i} g_{i}\mathcal{L}(\hat{y}, O_{i})
\label{eq:specloss}
\end{equation}

The gradients to each expert's network are therefore only dependent on the expert's own prediction, and not on other experts' predictions as with the cooperation loss. As a result, the specification loss allows each expert to learn independent from other experts, since other experts' predictions do not affect its gradients.

\dynamoe supports both the cooperation and the specification loss. While the cooperation loss is more common and easier to add to existing frameworks, we find that the specification loss delivers better results. Appendix \ref{sec:app-specloss} describes our implementation of the specification loss.


\citet{hierarch} first propose {\em hierarchical} MoEs, where each expert's network can also be an MoE itself (this can be continued recursively). Using a hierarchical MoE architecture reduces the branching factor of each gating network, which is particularly desirable for models with a large number of experts.

\citet{sgmoe} introduce sparse {\em top-$k$} gating, where only those $k$ experts predict on a sample, to whom the gating network assigned the highest weights. $k$ therefore serves as a hyperparameter that determines the amount of computation that is performed for each sample and allows the number of FLOPs (determined by $k$) to scale independent from the number of parameters (determined by the total number of experts $n$). Algorithm \ref{alg:moefwd} shows pseudo-code for the forward pass when using sparse top-$k$ gating.


\begin{algorithm}
\KwIn{\hspace{.77em}$x$: \; input sample\\
\hspace{3.8em}$k$: \; the number of experts to be selected\\
\hspace{3.8em}$G$: \; Gating network\\
\hspace{3.8em}$\{E_1,...,E_n\}$: \; A pool of $n$ experts}
\KwOut{\hspace{.06em}$y$: \; MoE prediction for $x$}
\vspace{.5em}
\vspace{.5em}
\begin{algorithmic}[1]
    \STATE{score $\leftarrow$ $G(x)$}
    \STATE{indices $\leftarrow$ argmax$_k$(score)}
    \STATE{($w_1$, ..., $w_k$) $\leftarrow$ normalize(\,score[indices]\,)}
    \STATE{$y \leftarrow$ 0 vector in output shape}
    \FOR{$i \leftarrow$ 1, ..., $k$}
    \STATE{$e \leftarrow$ indices[$i$]}
    \STATE{$y_i \leftarrow E_e(x)$}
    \STATE{$y \mathrel{+}= w_i * y_i$}
    \ENDFOR
    \STATE{\textbf{return} $y$}
\end{algorithmic}
\caption{MoE forward pass with top-k gating}
\label{alg:moefwd}
\end{algorithm}

Sparse gating introduces the problem that different experts typically receive different amounts of training data. In the beginning of the training, some experts receive more samples than others due to random initialization. These experts then have more training data than the others and thus deliver better predictions, which results in the gating network assigning them even more samples. This feedback loop can lead to all samples being assigned to very few (often only $k$) experts. To avoid such situations, various work suggests to add an imbalance penalty to the loss function \cite{sgmoe, switchtrans}. An example of a balance term is given in equation \ref{eq:balterm}.

\subsection{Systems work}
GShard~\cite{gshard} implements a distributed version of sparsely gated MoEs and scales MoE training to 600 billion parameters. The system demonstrates that MoEs can be trained efficiently on 2048 TPUs in parallel.

These results were further improved by Switch Transformer, which simplifies the MoE algorithm and scales to 1.6 trillion parameters \cite{switchtrans}. Switch transformer achieves the state-of-the-art predictive performance on several language tasks. The simplifications proposed in Switch Transformer include assigning each sample to a single expert (i.e., $k=1$) and using a simplified balance term given by Equation \ref{eq:balterm}. We adopt this balance term in \dynamoe.

\begin{equation}
    \mathcal{B} = \lambda*n*\sum_{i=1}^n T_i\cdot G_i
\label{eq:balterm}
\end{equation}

where $n$ is the total number of experts, $\lambda$ is a weighting factor of the regularizer, $T_i$ is the fraction of tokens assigned to the $i$-th expert, and $G_i$ is the fraction of gating network probability assigned to the $i$-th expert over the entire batch.

GShard and the system used to train Switch Transformer are not publicly available. There exist several open-source implementations using TensorFlow and PyTorch, but these implementations suffer from poor performance as they do not make any modifications to the frameworks to support MoEs natively. The implementations furthermore only work for a single GPU \cite{pytorchimpl-1, pytorchimpl-2}.
 
Tensor2Tensor \cite{t2t} provides an implementation of an MoE Transformer built on top of TensorFlow, but Tensor2Tensor has been deprecated and the MoE model does currently not function as intended.

FastMoE~\cite{fastmoe} is a distributed MoE training system based on PyTorch with support for multiple GPUs. However, FastMoE lacks crucial features such as support for capacity factors different from 1.0.

None of the above work, including GShard and Switch Transformer, addresses the dynamic behaviour of MoEs, which leads to an inefficient use of computational resources.

\begin{figure*}[t]
    \centering
    \includegraphics[width=\textwidth]{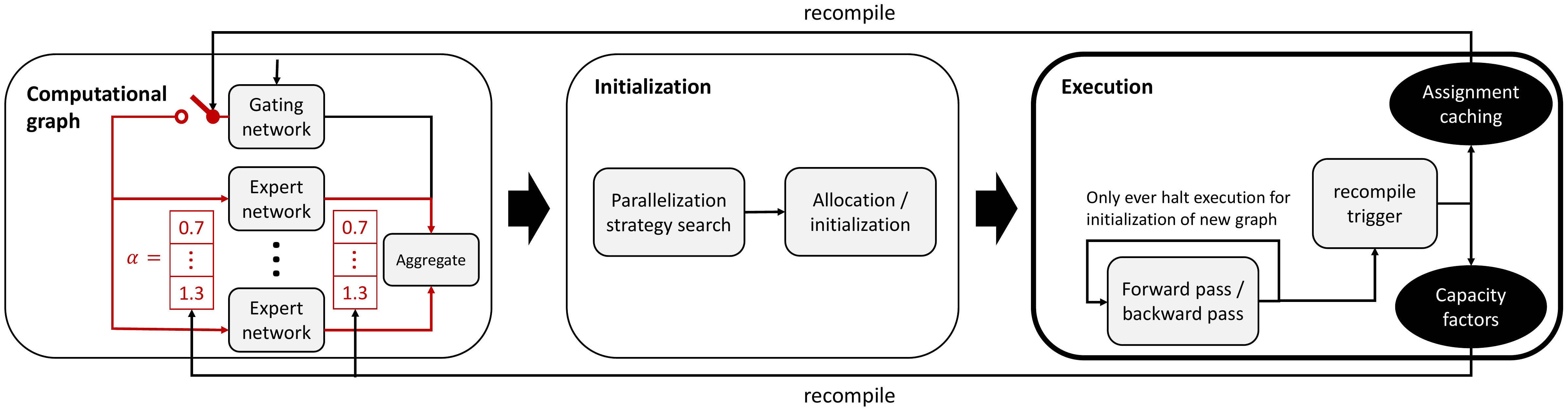}
    \caption{Overview of \dynamoeNoS, an MoE system the uses \recompile to leverage sample assignment caching and dynamic capacity factors for speed up, reduced memory pressure, and higher model quality.}
    \label{fig:overview}
\end{figure*}

\section{\recompile}
\label{sec:recompile-overview}
This section describes how \recompile works from the user's perspective. Our implementation of \recompile on top of FlexFlow is described in Appendix \ref{sec:rec-impl}. Additionally, we show in Appendix \ref{sec:rec-other} that \recompile can be applied to any other ML framework as well. Appendix \ref{sec:rec-other} also shows how the \recompile requirements are currently not fulfilled by popular ML frameworks such as PyTorch and TensorFlow and how these frameworks can thus benefit from implementing a mechanism like \recompile.

While \recompile is not specific to MoEs and can also be used to optimize other models, this paper focuses on using \recompile for MoEs. After this section's general introduction to \recompile, Section \ref{sec:recompile} presents two MoE-specific optimizations that are enabled through \recompile. 

Conceptually, \recompile comprises of a mechanism that compiles the computation graph again from scratch but only reallocates resources where required. For example, when changing the batch dimension of a tensor, any memory region containing weights remains untouched. This allows \recompile to avoid prohibitive overheads that would result from naively recompiling the entire graph.

\recompile further allows to plan such recompilations while preventing overheads from the graph adjustment decisions. This is important because computing when and how to alter the computation graph can be expensive and needs to be carried out frequently (i.e. after each mini-batch). In fact, determining how to alter the computation graph could for example require the prediction of an additional neural network, as discussed in Section \ref{sec:conclusion}. To avoid adding such overheads to the overall training/inference runtime, \recompile runs graph adjustment decisions on CPU while the training or inference of the model continues on GPU.

In summary, an implementation of \recompile must satisfy the following three requirements.


\begin{enumerate}
    \item Users can monitor \textit{model metrics} with negligible overhead. A model metric can be anything that informs the user about the current state of the model during inference or training, such as training loss.
    \item Users can alter the computation graph with little overhead. When and how the graph is altered is determined by a user-provided \textit{recompile trigger}, which may base its decision on model metrics.
    \item Executing the recompile trigger and computing the hardware mapping of the new graph is overlapped with inference/training using the current graph. For example: A user can conduct an expensive computation to decide how to alter the graph or conduct a long search for a new parallelization strategy while model training continues in parallel.
\end{enumerate}

By implementing a \recompile mechanism, \dynamoe therefore allows users to easily alter the computation graph based on arbitrary, user-defined events without incurring any scheduling overheads. \dynamoe further exposes operators and dynamically-sized tensors that leverage \recompile and can be used like any other operators or tensors in only one line of code.





\section{\recompileNoS-Enabled Optimizations}
\label{sec:recompile}


This section presents two optimizations for MoE that are naturally enabled through \recompile and only require few lines of user code. These optimizations are also schematically depicted in Figure \ref{fig:overview}.

\subsection{Dynamic capacity factors}
\label{subsec:dynamic_capacity}

Over the course of MoE training, the gating network may assign varying numbers of samples to an expert. Current MoE frameworks 
require the declaration of fixed-sized tensors when constructing an MoE model. To deal with this, users currently declare overly big tensors (i.e. increase the batch dimension) and {\em{drop}} samples that still don't fit into the tensors (i.e. ignore them during back propagation). As a result, when a mini-batch contains fewer samples than the declared capacity for an expert, the first entries of the input tensor will be used for these samples and the remaining entries will be unused during training by setting gradients to zeroes.

\citet{switchtrans} denotes the \textit{expert capacity} to be the number of samples that an expert's input tensor can hold.  For an MoE model with $n$ experts, where $k$ experts are assigned to each sample, an {\em expert capacity factor} $\alpha$ is used to compute each expert's capacity $\mathcal{C}$ as shown in  Equation \ref{eq:expert_cap}.

\begin{equation}
    \mathcal{C} = \alpha*batch\_size*k/n
    \label{eq:expert_cap}
\end{equation}

Note that a high expert capacity factor requires additional computation (FLOPs), memory and data transfer. On the other hand, low expert capacity factors lead to more samples being dropped which leads to a lower model quality. Defining the expert capacity factor thus involves a trade off between model performance and computational resources.

Using \recompile, we can largely improve this trade off by using \textit{dynamic capacity factors} that adapt according to the capacity that an expert actually needs. To do so, \dynamoe measures how many samples the gating network assigns to each expert and creates a model metric for that. As the expert's capacity requirement changes over the course of training, the expert's capacity factor can be adjusted accordingly by calling \recompileNoS, which allows users to both adjust the capacity dynamically over time and tailor it separately to each expert.

\subsection{Sample Assignment Caching}
\label{sec:meth-cachescore}

During the training of MoEs, it is critical that experts specialize to subdomains of the input --- otherwise, the increase in parameter size is not as effective since experts are redundant in functionality and most parameters have the same, general-purpose function. For an expert to specialize to a subdomain, the expert must be trained on samples that predominantly are elements of that subdomain. If too many samples from the remaining input domain are assigned to the expert, the expert will not specialize and will instead attempt to reduce the loss for predictions on the entire input domain. 

We hypothesize that the sample assignments of the gating network converge early in training. Once the gating network has decided to assign a certain subdomain to some expert, this decision is reinforced because the expert specializes on that subdomain, thus delivering better performance on it than other experts and incentivizing the gating network to keep assigning samples of that subdomain to the expert. Once this feedback loop begins, the gating network will mostly fine tune the weighting of experts and will only reassign few samples.

\begin{figure}[h]
    \centering
    \includegraphics[width=0.37\textwidth]{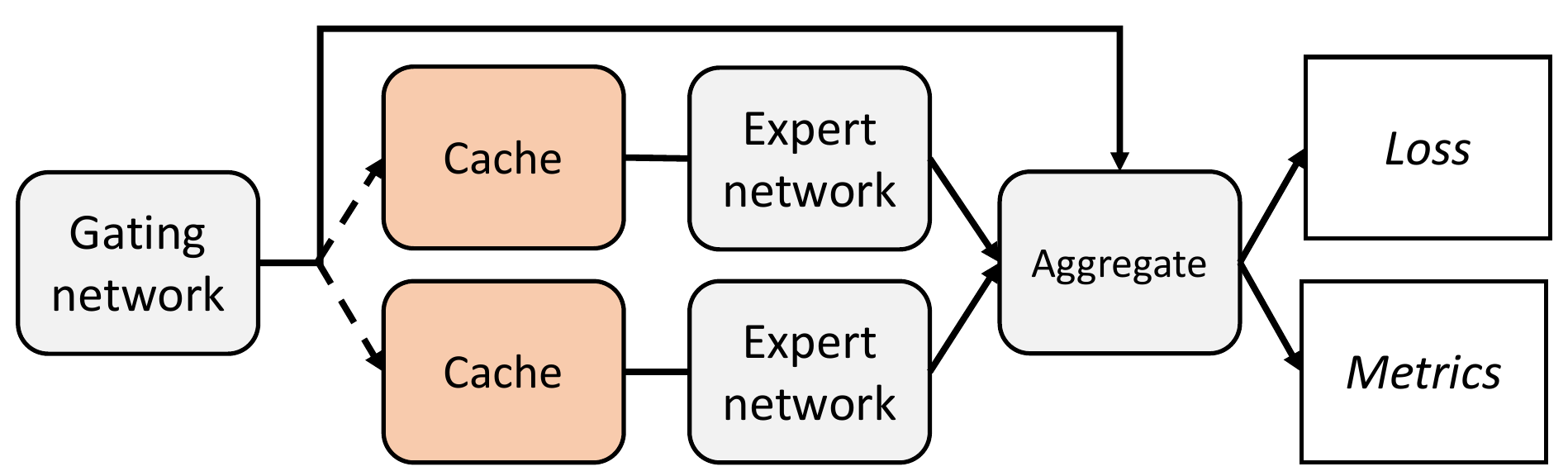}
    \cprotect\caption{The \verb|Cache| operator eliminates the dashed dependency after the sample assignments have converged.}
    \label{fig:cache-graph}
\end{figure}

Early convergence on sample assignments can be leveraged to achieve a higher degree of model parallelism. Since expert batches can only be formed once the sample assignments are known, the forward passes of the experts can only be executed once the forward pass of the gating network has finished. However, if we know the sample assignments before the gating network has finished its forward pass, this dependency can be eliminated and the gating network and experts can execute their forward passes in parallel. Figure \ref{fig:cache-graph} shows the ``disposable'' dependency as dashed arrows.

Using \recompile, we can detect when the sample assignments have converged and eliminate the dependency thereafter. More specifically, we can create a model metric on how many samples of a mini-batch are reassigned to different experts between two consecutive training epochs. If only a few samples are reassigned, we can cache the sample assignments and use the cached assignments to create expert batches before the gating network has finished its forward pass.

\dynamoe provides a \verb|Cache| operator that remembers sample assignments and creates a metric recording how many samples have been reassigned between two consecutive epochs. The \verb|Cache| operator is shown in red in Figure \ref{fig:cache-graph}. Even when executing expert and gating net forward passes in parallel, the \verb|Cache| operator will update the cached sample assignments after the gating network has finished its forward pass. However, these new assignments will only be used in the next epoch. Thus, after detecting that the sample assignments have converged, the model will use the previous epoch's sample assignments to achieve a higher degree of parallelism.

\begin{figure}[h]
    \centering
    \includegraphics[width=0.48\textwidth]{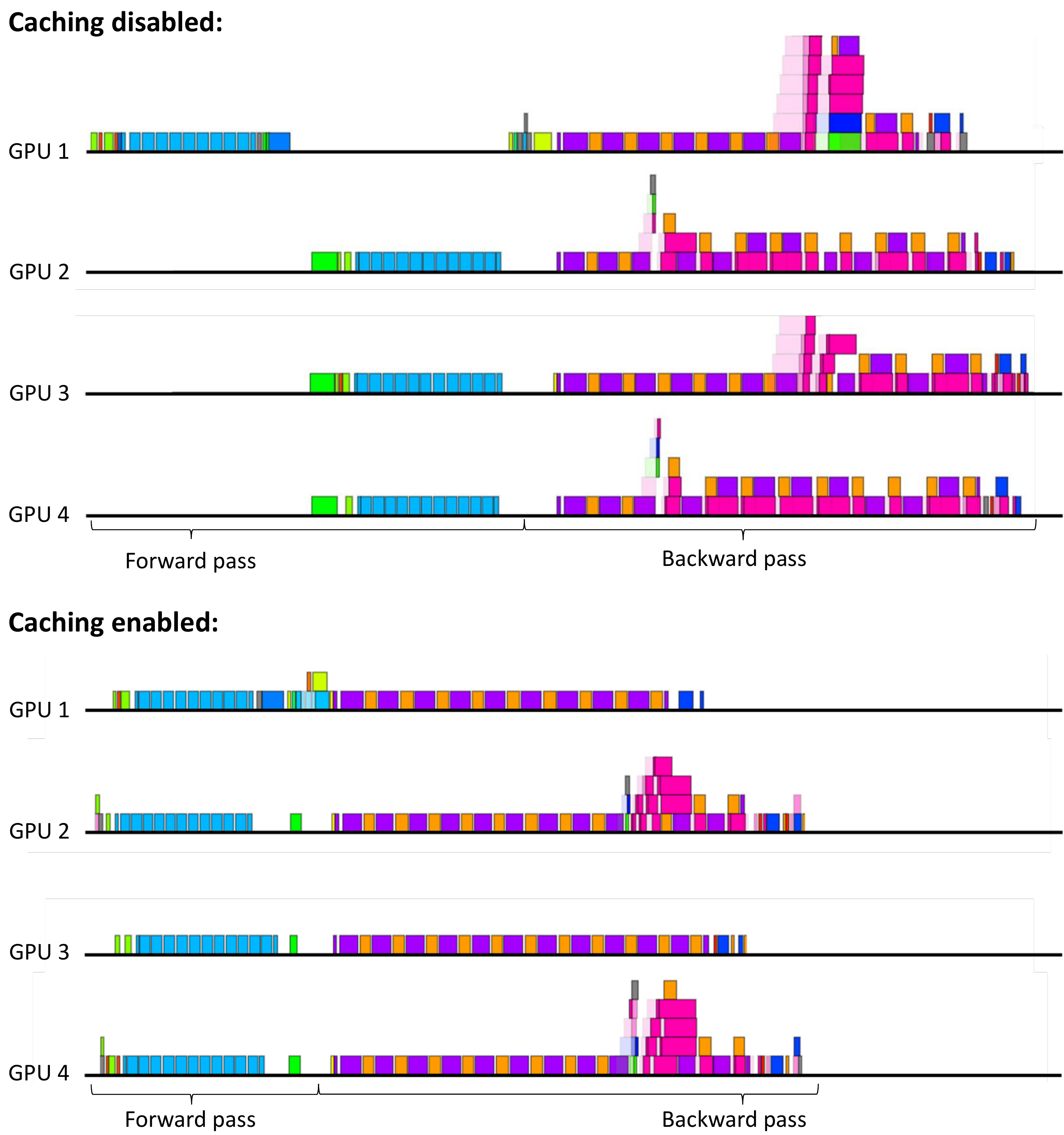}
    \cprotect\caption{GPU 1 executes the gating network, GPUs 2-4 execute one expert each. Each box represents the execution of a task. Caching allows to execute the expert forward passes in parallel to the gating net forward pass.}
    \label{fig:cacheprof}
\end{figure}

Figure \ref{fig:cacheprof} shows how tasks are scheduled with and without caching enabled on four GPUs. Note how the forward passes of the gating network and the experts happen in parallel when caching is enabled.

\section{Evaluation}
\label{sec:results}
We evaluate \dynamoeNoS in three aspects. Section~\ref{sec:res-comb} compares the end-to-end MoE training performance between \dynamoe and existing MoE systems; Section \ref{sec:res-capfac} evaluates dynamic capacity factors; and Section \ref{sec:res-caching} evaluates sample assignment caching. We hereby focus on evaluating individual MoE layers to reduce the influence of further operators on runtime as much as possible. 

All experiments were conducted on a machine with four NVIDIA Tesla P100 GPUs and two Intel Xeon E5-2640 v4 CPUs. Unless otherwise stated, all experiments are conducted on CIFAR10 \cite{cifar10} using CNNs \cite{cnn} as gating and expert modules.

\subsection{End-to-end Performance}
\label{sec:res-comb}
\begin{figure}[h]
    \centering
    \includegraphics[width=0.48\textwidth]{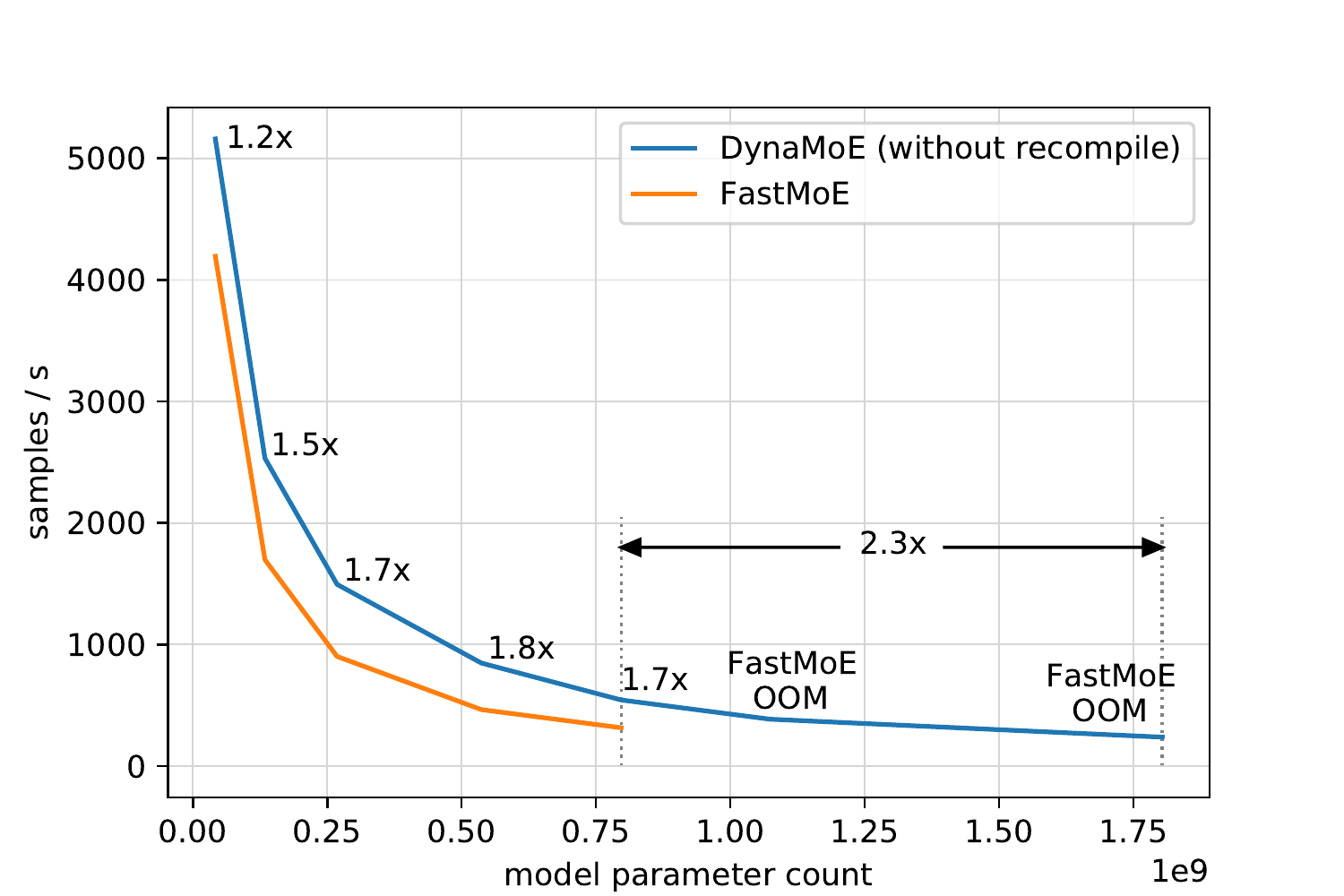}
    \caption{\dynamoe without \recompile optimizations is up to $1.8\times$ faster than FastMoE. Furthermore, \dynamoe is able to support $2.3\times$ bigger models before running out of memory (OOM).}
    \label{fig:fastmoe}
\end{figure}

Figure \ref{fig:fastmoe} compares \dynamoe against FastMoE, a state-of-the-art open-source MoE implementation \cite{fastmoe}. For the comparison, one CPU and one GPU were used. No \recompile optimizations are enabled and the blue line reflects the baseline \dynamoe performance.

The benchmark uses the same model architecture that FastMoE uses in their benchmarking script on GitHub (multilayer perceptron modules). We recreated this architecture in \dynamoe and also set all further hyperparameters equal. As in FastMoE's benchmarking script, we used a fixed, randomly generated data batch and measure the end-to-end throughput for training over 10,000 steps. The scaling of the parameter size (x-axis) is done by increasing the number of experts $n$ in the network.

Figure \ref{fig:fastmoe} shows that \dynamoe yields a speed up of up to $1.8\times$ over FastMoE on the larger model sizes, which are the ones that matter most given the goal of MoEs is to scale model parameters. \dynamoe was furthermore able to support $2.3\times$ larger models before running out of memory (OOM). We speculate that part of the speed up comes from \dynamoe using a static computation graph and implementing corresponding optimizations. FastMoE on the other hand sits on top of PyTorch and uses a dynamic computation graph, which does not allow for all of these optimizations.

\begin{figure}[h]
    \centering
    \includegraphics[width=0.48\textwidth]{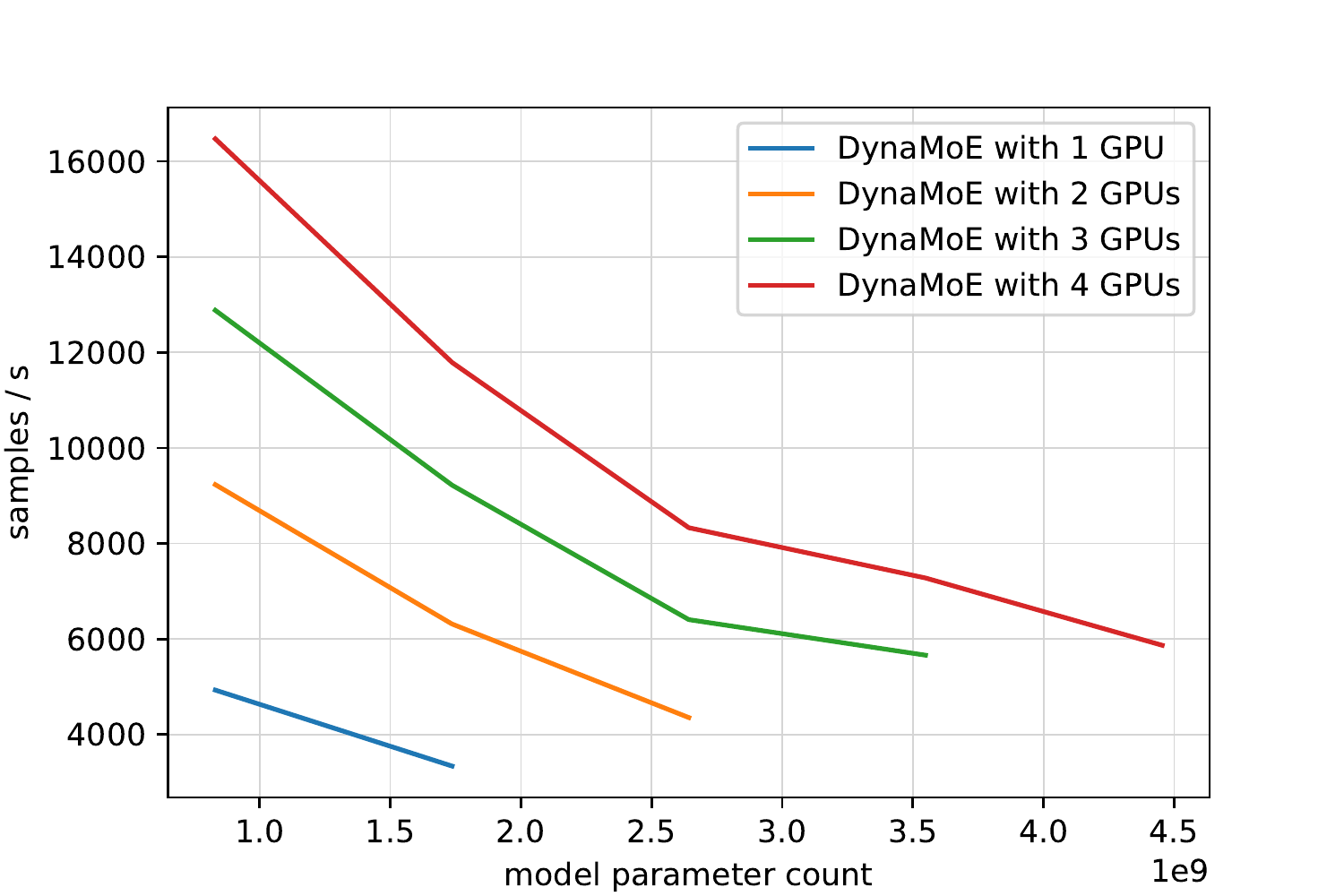}
    \caption{\dynamoe scales linearly in both the speed up and the supported model size when adding more GPUs.}
    \label{fig:scaleout}
\end{figure}

Figure \ref{fig:scaleout} further shows that \dynamoe scales linearly when adding GPUs, both in terms of throughput and model size that is supported before running out of memory.

\begin{figure}[h]
    \centering
    \includegraphics[width=0.48\textwidth]{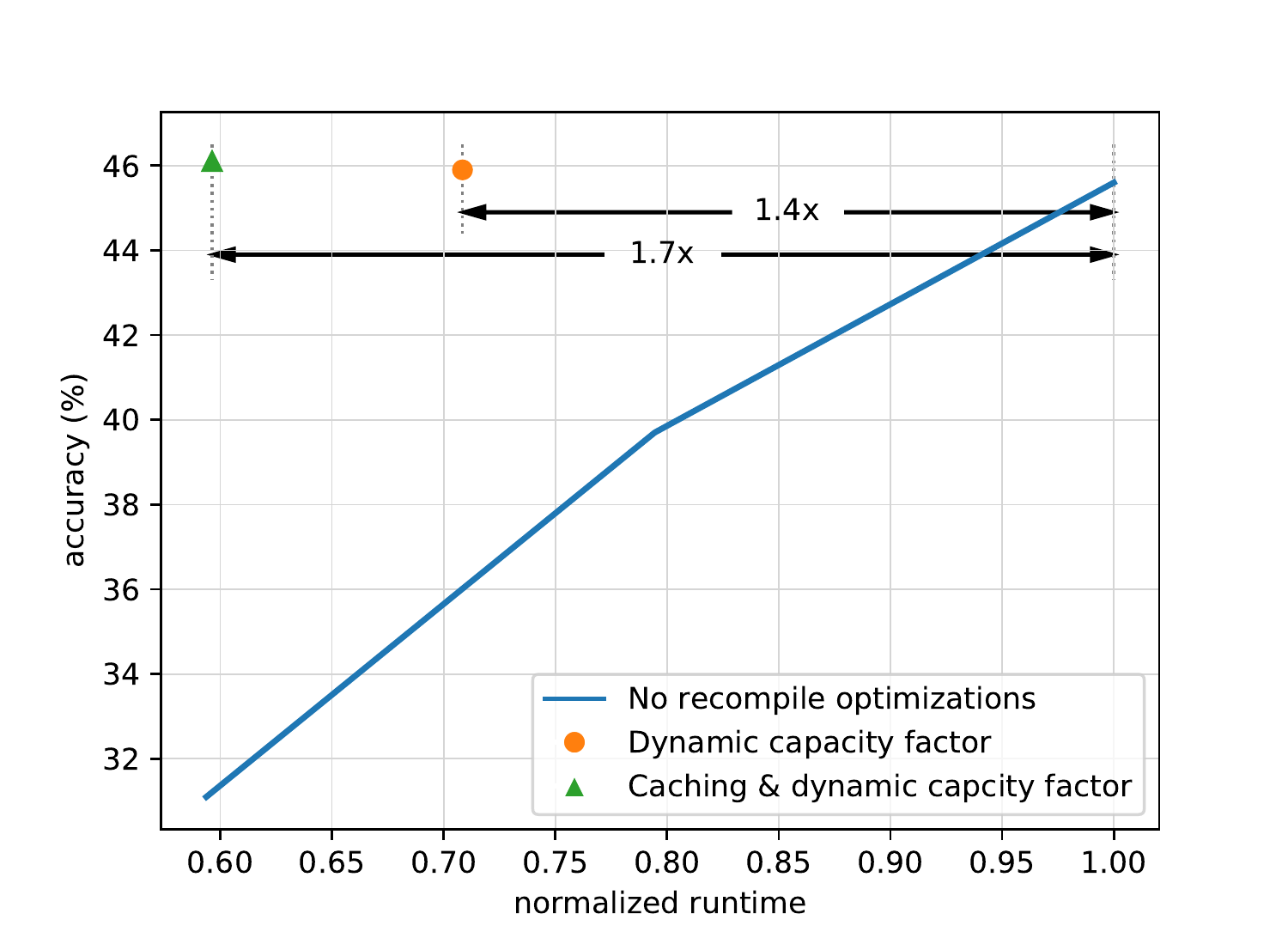}
    \caption{The combined \recompile optimizations achieve $1.7\times$ speed up while not affecting the accuracy.}
    \label{fig:run-acc-gen}
\end{figure}

We now compare the \recompile optimizations against \dynamoe with varying static capacity factors. Since FastMoE does not support capacity factors different from 1.0, we omit FastMoE in Figure \ref{fig:run-acc-gen}. 


Figure \ref{fig:run-acc-gen} shows how the dynamic capacity factors achieve the same accuracy as when using a static capacity factor that ensures that no samples are dropped (in this case, $\alpha = 7.0$). The dynamic capacity factor hereby yields a speed up of $1.4\times$. When sample assignment caching is also enabled, the speed up increases to a total of $1.7\times$ while still achieving the same accuracy. Using both optimizations, we get the same throughput as a static capacity factor of $\alpha = 1.0$ but without the losses in model quality. 

Using only the caching optimization with static capacity factors doesn't yield a speed up in our experiments, because the experimental setup did not allow for a hardware mapping that exploits additional model parallelism. The hardware mappings considered were limited to expert parallelism, which means that the gating network and any expert module must entirely be placed onto one device and a single module cannot be distributed among several devices. When relieving this constraint, or choosing a different experimental set up (i.e. different hyperparameters or different device topology), caching can also yield speed ups with static capacity factors, as we show in Section~\ref{sec:res-caching}.


Finally, we compare the cooperation loss with the specification loss and also look at the baseline of keeping the expert assignments uniformly fixed, as this hints at where the better performance is coming from: Better expert assignments or better experts.

\begin{figure}[h]
    \centering
    \includegraphics[width=0.46\textwidth]{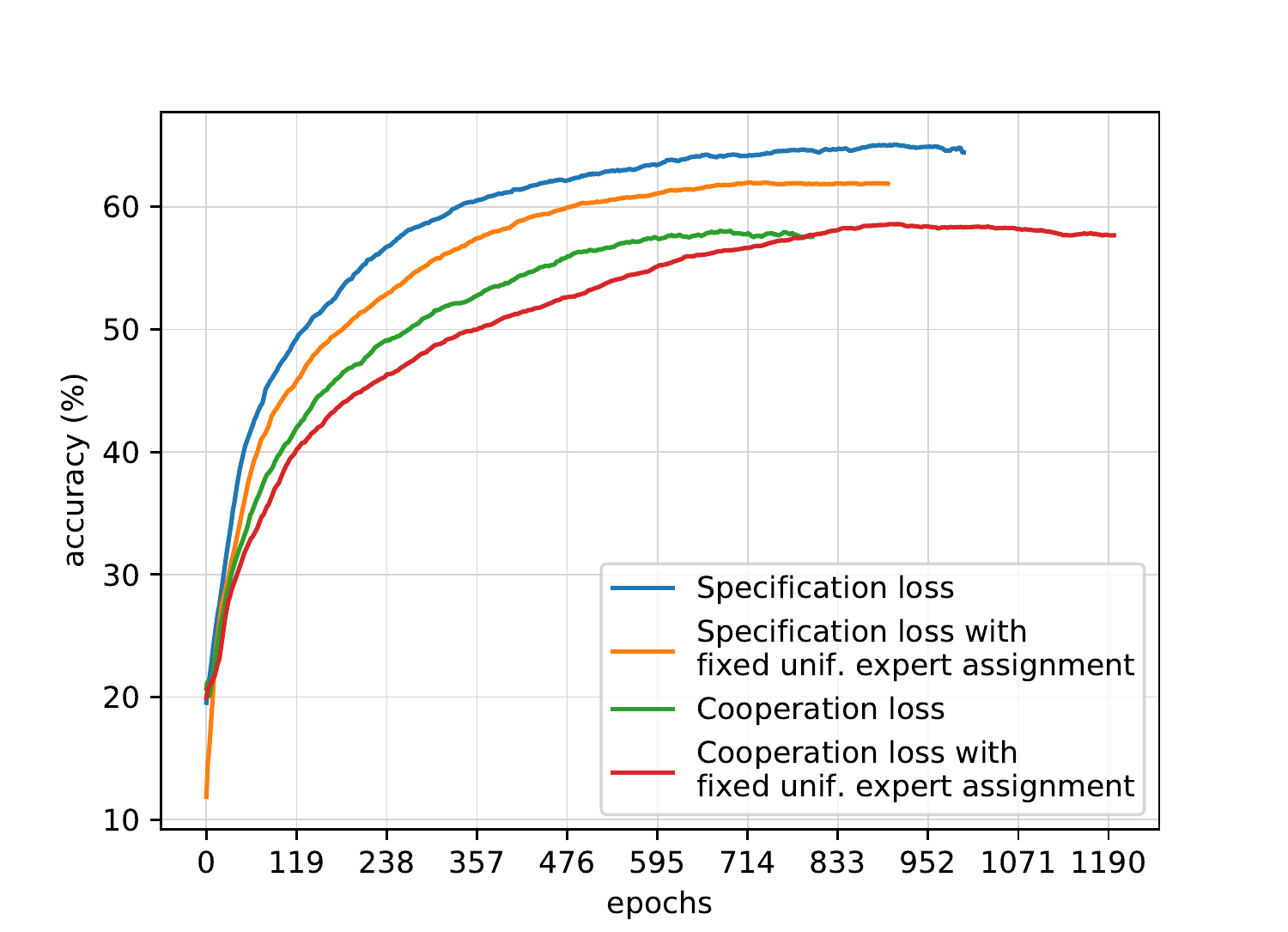}
    \caption{The specification loss significantly outperforms the cooperation loss.}
    \label{fig:spec-coop}
\end{figure}

Figure \ref{fig:spec-coop} shows that the specification loss significantly outperforms the cooperation loss. Furthermore, the specification loss seems to be the better loss for both the experts and the gating network by themselves: The specification loss outperforms the cooperation loss when keeping the sample assignments fixed to a uniform random assignment, which suggests that the specification loss is better for expert training independent of the gating network. Furthermore, using learned sample assignments had a greater impact when using  specification loss, which suggests that the specification loss is also a better loss function for the gating network.\footnote{It is unexpected that for cooperation loss, learned gating net assignments did not outperform uniform random assignments.}

\subsection{Dynamic Capacity Factors}
\label{sec:res-capfac}
\begin{figure}[h]
    \centering
    \includegraphics[width=0.46\textwidth]{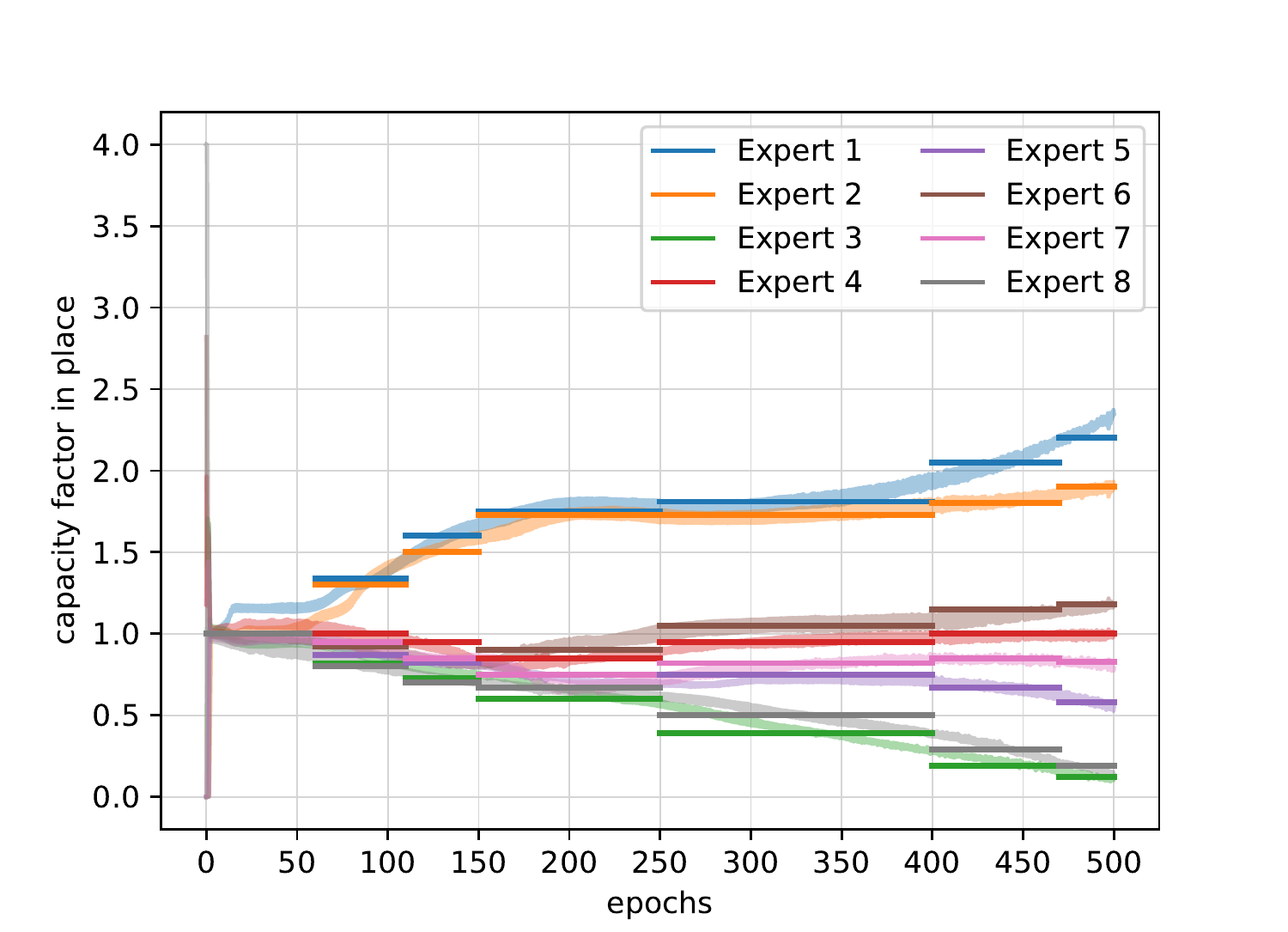}
    \caption{The capacity factors dynamically adjust to the actual capacity requirements of the experts.}
    \label{fig:cap-use}
\end{figure}

The lines with a dim color in Figure \ref{fig:cap-use} show the ratio between the number of samples that are assigned to each expert and the average number of samples assigned to an expert. For example, if expert $E$ has a ratio of 2 at a given point in training, it means that the gating network assigned twice the average number of samples to expert $E$ at that point in time.

Figure \ref{fig:cap-use} shows how the number of samples assigned to an expert varies over the course of training and is imbalanced. As discussed in Section~\ref{sec:recompile}, a common way to deal with  imbalance is to determine a capacity factor that multiplies each expert's capacity, but bigger capacity factors hereby introduce a larger overhead in memory usage, runtime and inter-device data transfer. However, if an expert doesn't have enough capacity, samples are ignored during training, which affects model quality. 

In previous work, the capacity factor is determined statically (fixed over the course of training) and globally (the same for all experts). However, static, global capacity factors can neither capture the variations in imbalance over the course of training, nor the variations between individual experts. In contrast to static, global capacity factors, Figure \ref{fig:cap-use} shows how \dynamoe uses different capacity factors for different experts and also changes the capacity factors over the course of training. The lines in bright color show the capacity factors that \dynamoe used for a given expert at a given point in training.

\begin{figure}[h]
    \centering
    \includegraphics[width=0.46\textwidth]{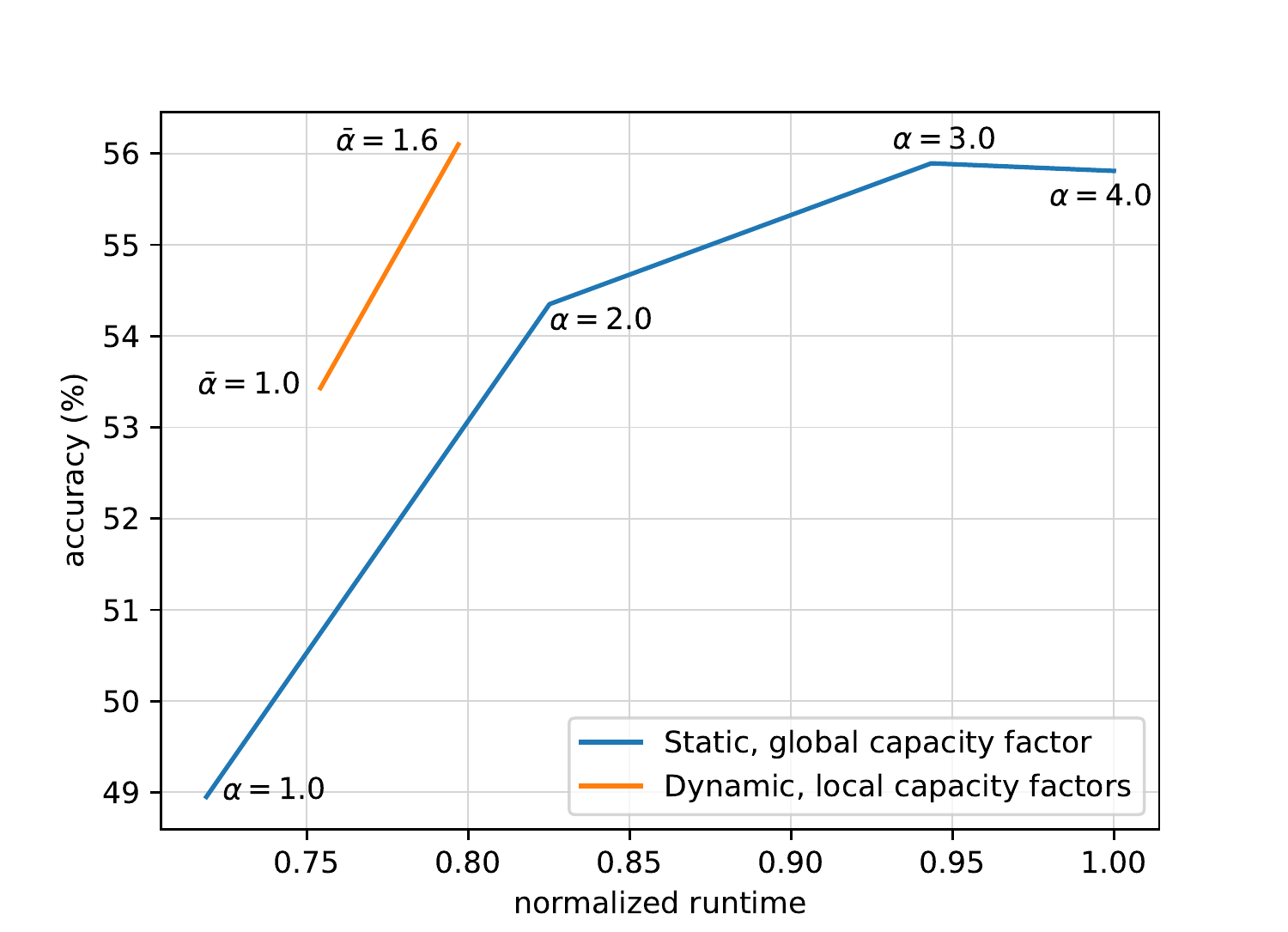}
    \caption{Dynamic, local capacity factors offer a better runtime-accuracy trade off than static, global ones.}
    \label{fig:capfac-tradeoff}
\end{figure}

On average across all experts, the needed capacity factor to not drop samples is always 1.0 (by definition). However, creating a \recompile policy with an average capacity factor of 1.0 and no sample drops is infeasible: Since there are slight variations in the number of samples assigned to an expert between each training iteration, this policy would need to recompile every single iteration. Since this would eliminate any speed up, designing a \recompile policy with small average capacity factor, few sample drops and a tolerable number of recompilations is non-trivial. It helps however, to allow the average capacity factor to be above 1.0. Thus the runtime-accuracy trade off is also present when using dynamic capacity factors. Figure \ref{fig:capfac-tradeoff} shows that \recompile policies with different average dynamic capacity factors offer a more attractive trade off in comparison to using static capacity factors.



\subsection{Sample Assignment Caching}
\label{sec:res-caching}
\begin{figure}[h]
    \centering
    \includegraphics[width=0.46\textwidth]{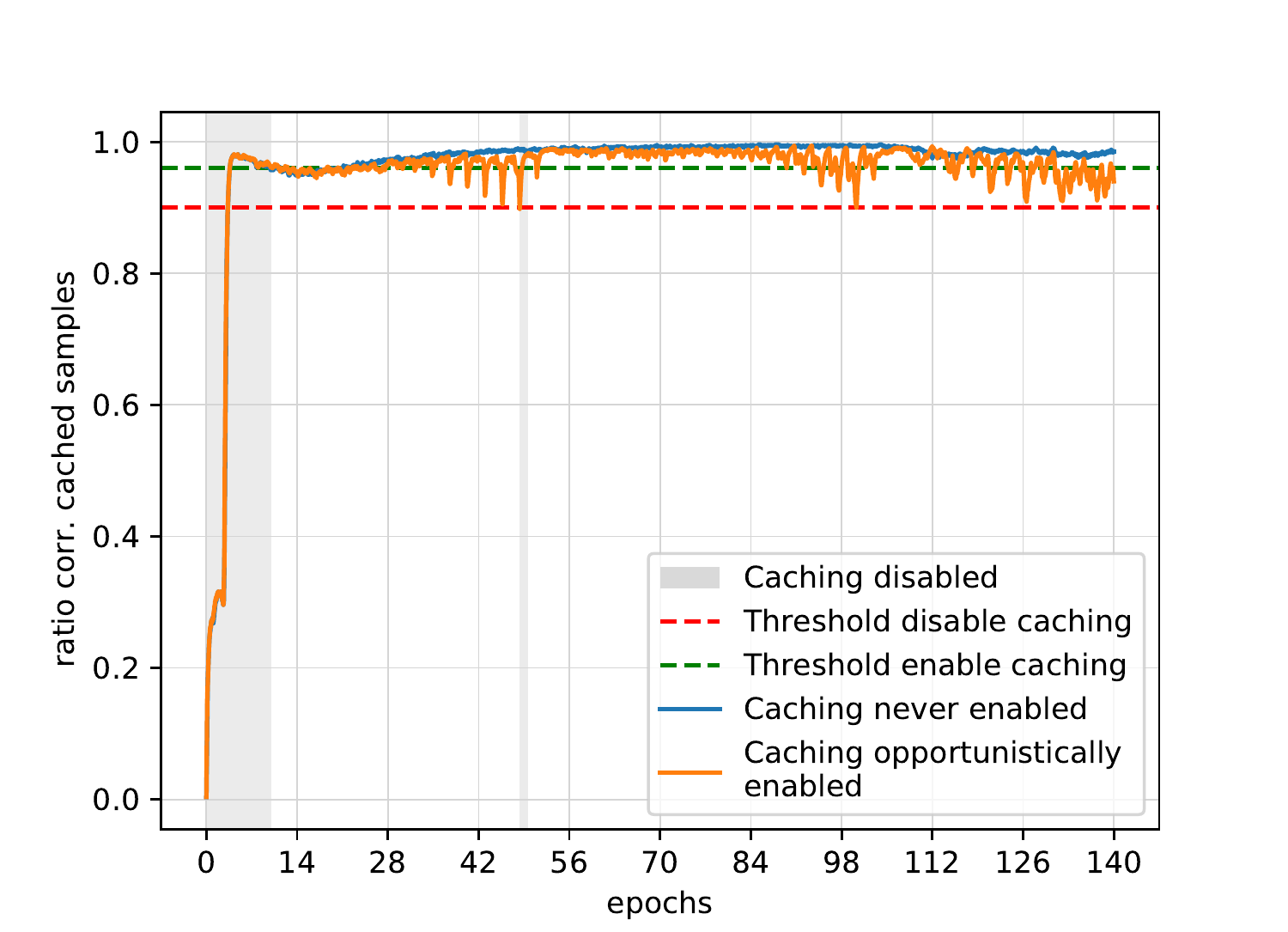}
    \caption{During phases with few samples being reassigned, the system switches to using cached sample assignments for speed up. The system will further detect phases with much reassignment and switch back to not using cached assignments.}
    \label{fig:cache_score}
\end{figure}

Figure \ref{fig:cache_score} shows that sample assignments do in fact converge early in training. The y-axis depicts the fraction of samples that have not been reassigned by the gating network between two consecutive epochs, which is the fraction of samples that have been correctly cached. In the experiment conducted in Figure \ref{fig:cache_score}, we switched sample assignment caching on if at least 96\% of the samples were correctly cached; caching is switched off if the percentage of correctly cached samples drops below 90\%, which happened for a very short period during epoch 48. We additionally did not switch caching on before epoch 10. Using this policy, caching yielded a speedup of $1.15\times$.

We have observed early sample assignment convergence as in Figure \ref{fig:cache_score} on different data sets and using different architectures and hyperparameters. However, even if sample assignments should not converge early, having the caching optimization disabled did not have a significant impact on performance in comparison to not running any caching related code at all.

Furthermore, we observed that the final accuracy as well as the accuracy progression over the course of training is generally unaffected by the caching optimization. For the training depicted in Figure \ref{fig:cache_score}, Appendix \ref{sec:cache_acc} shows the model's test accuracy progression over time as an example of how little the caching optimization affects the progression.


\section{Conclusion \& Future Work}
\label{sec:conclusion}
Using the example of Mixture of Experts, we demonstrate how \recompile allows for more efficient use of computational resources for models with dynamic behaviour. While the MoE algorithm specifically aims at reducing computational cost when scaling models, current implementations often include overheads of $2\times$ in terms of FLOPs and a significantly larger memory usage just to deal with the dynamic data flow in MoEs. \recompile hereby not only saves time and memory but also improves the model quality because fewer samples are dropped during training and because models can be scaled even further due to the more efficient resource use.

Although we have successfully run \dynamoe on multiple nodes, scaling our system to huge sizes consisting of several 100s of GPUs remains to be tried out.
There also remain many future research questions regarding MoEs in general as well as \recompile in particular. Some questions related to \recompile are outlined here:

\begin{itemize}
    \item \textbf{Graph adjustments:} It's not trivial to determine when and how to alter the computation graph during training and inference. The policies used in our experiments are relatively naive and can neither guarantee to be close to some optimum, nor can they generalize well to other hyperparameter and data set choices. \recompile provides a framework where also expensive algorithms can serve as a \recompile policy without incurring overheads in the overall training time. 
    \item \textbf{Remapping onto hardware:} The computation graph might be adjusted so much over the course of training that remapping the graph using a different parallelization strategy can pay off. However, while graph adjustments are relatively cheap, remapping the graph can induce large overheads and should be planned carefully. A holistic solution for a \recompile policy should consider both, graph adjustments and hardware remappings, at the same time. The policy must hereby consider the expense of remapping the graph (which depends on how many graph components are mapped to a different device) as well as the likelihood of that mapping paying off over time.
    \item \textbf{Training instabilities:} Instabilities in training remain a problem of MoEs and advances in this direction also benefit the \recompile optimizations: During unstable phases in training, sample assignment caching is switched off and therefore doesn't yield any speed up. Furthermore, during times of large fluctuations in the expert capacities, dynamic capacity factors will frequently issue recompilations which could have a negative effect on the speed up in extreme cases. 
\end{itemize}

\clearpage




\begin{figure*}
    \centering
    \includegraphics[width=0.8\textwidth]{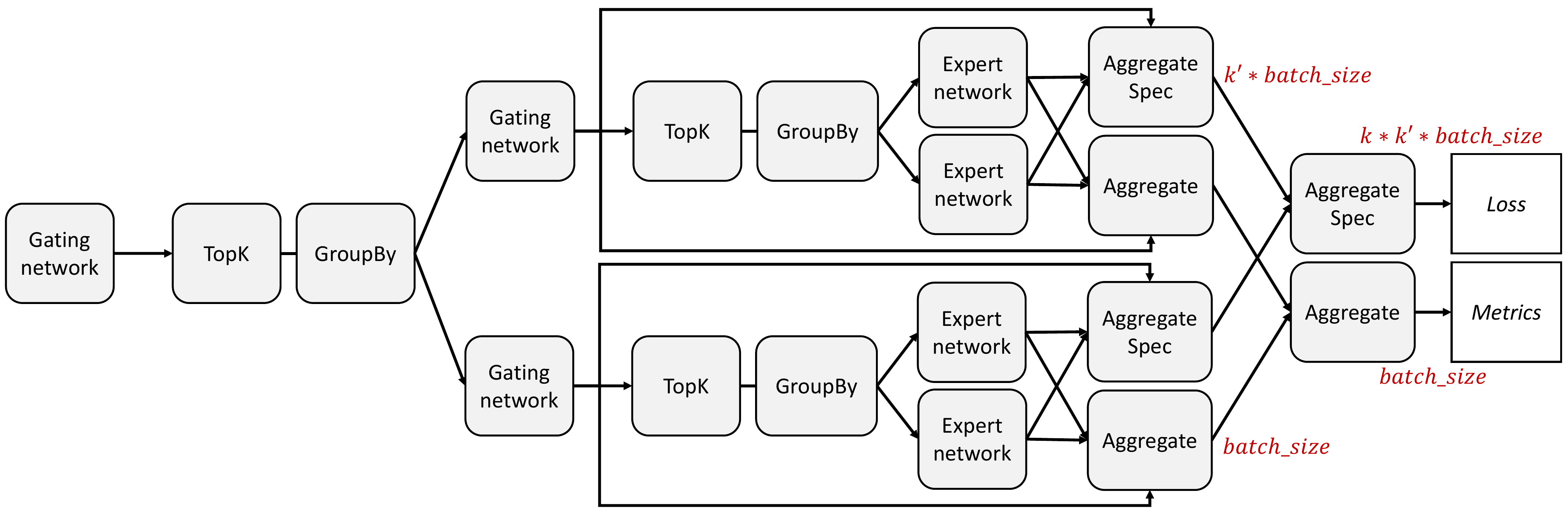}
    \caption{Hierarchical model with specification loss}
    \label{fig:graphhierarc}
\end{figure*}
\newpage
\appendix
\section{Specification loss as operator}
\label{sec:app-specloss}

The specification loss requires the prediction of each chosen expert and does not operate on the actual model output (see equation \ref{eq:specloss}). If $k$ experts are assigned to each sample, the specification loss thus requires $k*batch\_size$ expert predictions -- and not $batch\_size$ many. This makes it impossible to support the specification loss in the same way that other loss functions are supported.

When using the specification loss, the network output must consists of two tensors: The concatenated expert predictions for the loss computation, and the aggregated expert predictions as the actual model output (for example, to compute metrics such as accuracy on the model performance).

In \dynamoe, Mixutre of Experts models are defined using three MoE specific operators:
\begin{itemize}
    \item \textbf{TopK}: Get indices and values of the biggest $k$ values from a list of $n$ values. This is used to extract the chosen experts from the gating net logits. 
    \item \textbf{GroupBy}: Given a data batch and a mapping betwen samples and experts, \verb|GroupBy| groups the samples into $n$ tensors where the $i$-th tensor contains all samples that are mapped to the $i$-th expert. 
    \item \textbf{Aggregate}: Given the expert predictions and their respective weighting, create the weighted average over the predictions. This weighted average is the output prediction of the MoE.
\end{itemize}

Intuitively, supporting the specification loss can be achieved by altering the \verb|Aggregate| operator to take a Boolean $\beta$ that determines which loss is being used. If $\beta$ is true, the specification loss is used and \verb|Aggregate| has two outputs. Else, the cooperation loss is used and \verb|Aggregate| only has one output. Under the hood, this would mean that there are two different \verb|Aggregate| operators but this would be hidden to the user.

However, if the user wants to build a hierarchical model using the specification loss, \verb|Aggregate| would need to take two inputs, the concatenated expert predictions and the aggregated expert predictions. This would require another Boolean and would lead to four different \verb|Aggregate| operators under the hood. Furthermore, the API would get confusing as it is not self-explanatory how to define the two Booleans to express the wanted graph functionality. Instead, we expose a special \verb|AggregateSpec| operator to the user, which allows for a more intuitive graph construction.

\begin{figure}[h]
    \centering
    \includegraphics[width=0.48\textwidth]{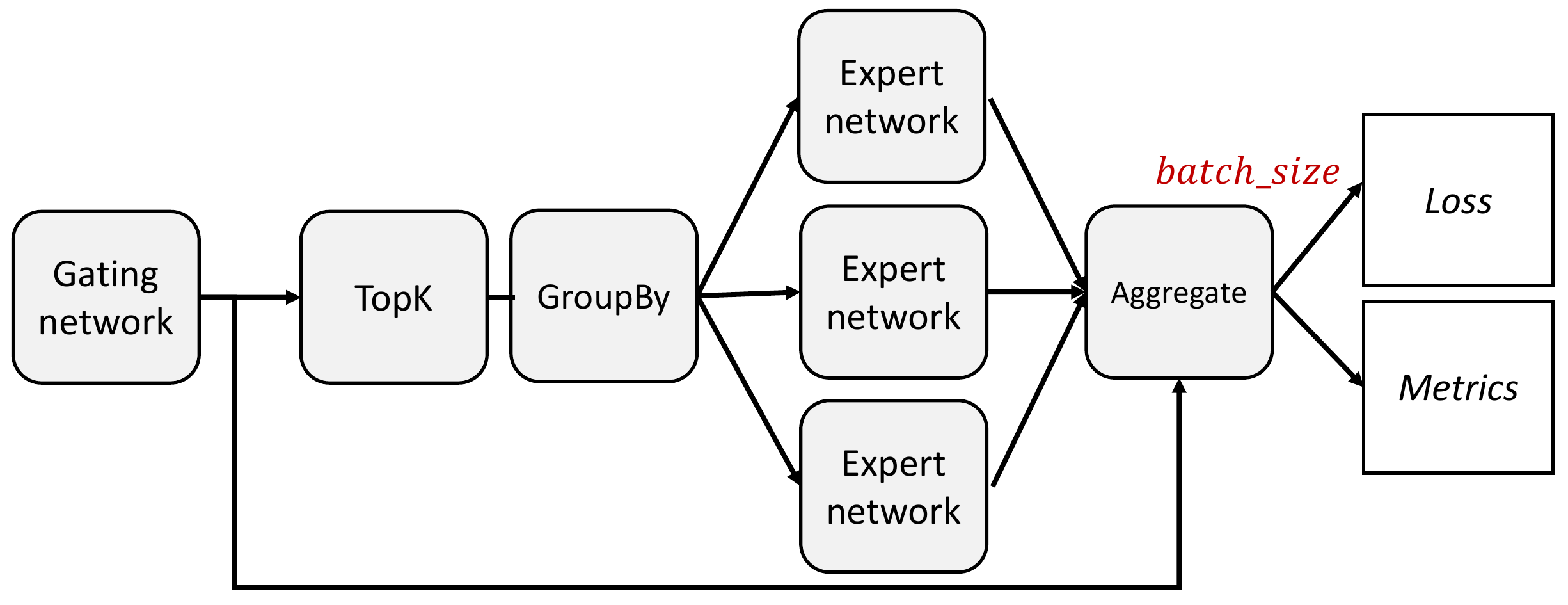}
    \caption{Cooperation loss}
    \label{fig:graphcoop}
\end{figure}

Both aggregate operators have a single output, where \verb|Aggregate| outputs the aggregated expert predictions and \verb|AggregateSpec| outputs the concatenated expert predictions. \verb|Aggregate|'s output makes up for the MoE's final output predictions --- that means that \verb|Aggregate|'s output is used as the inferred labels, or to compute the model's accuracy. When using the cooperation loss, \verb|Aggregate|'s output is also used for the loss computation (see Figure \ref{fig:graphcoop}).

\begin{figure}[h]
    \centering
    \includegraphics[width=0.48\textwidth]{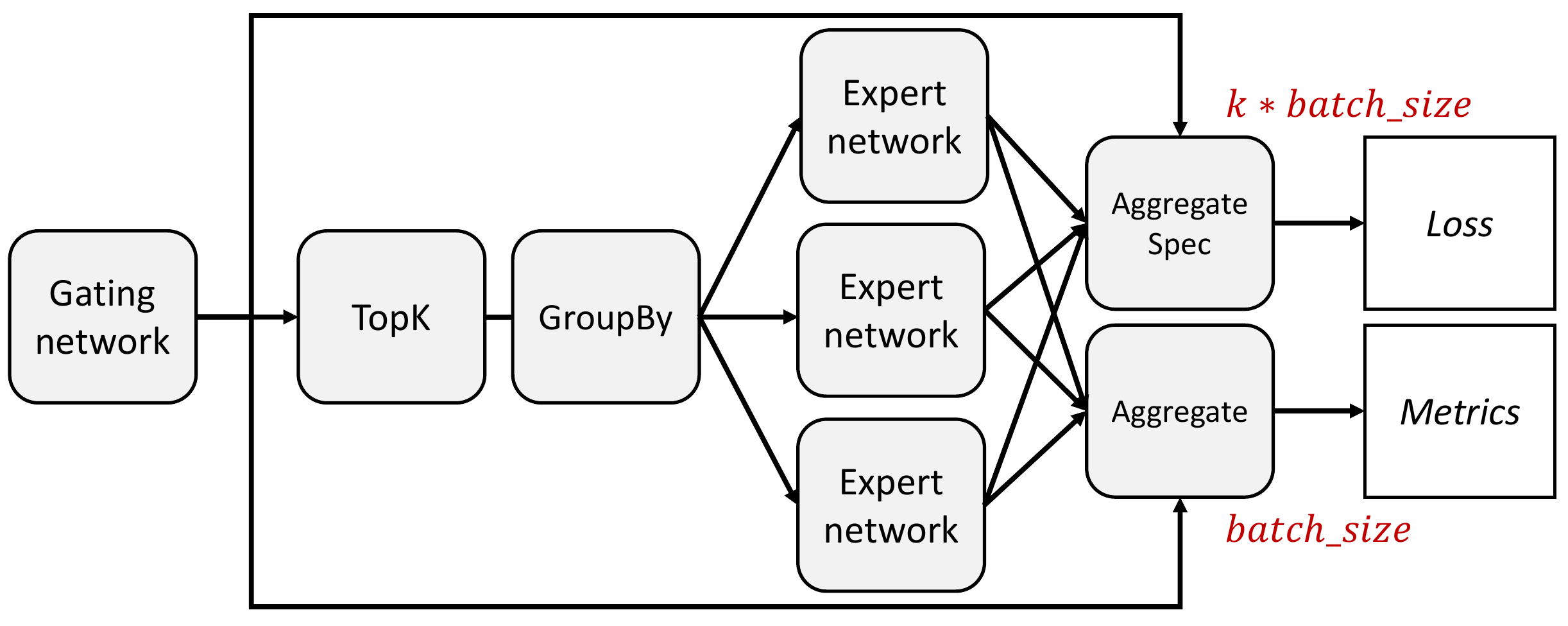}
    \caption{Specification loss}
    \label{fig:graphspec}
\end{figure}

When using the specification loss however, \verb|Aggregate|'s output is only used as the model output but is not used for the loss computation, since the loss computation requires all expert predicitions, i.e. \verb|AggregateSpec|'s output. \dynamoe lets users easily add an \verb|AggregateSpec| operator and specify that the \verb|AggregateSpec| operator's output should be used for the loss computation. This is depicted in Figure \ref{fig:graphspec}. Figure \ref{fig:graphhierarc} further depicts the graph of a hierarchical model -- note how losses can even be mixed in hierarchical models and the inner MoEs could use a cooperation loss while the outer ones use a specification loss.

In conclusion, by exposing an additional operator for the computation of the specification loss, we allow for flexibility (i.e. defining hierarchical models) while at the same time not introducing confusing Booleans that are not self-explanatory.


\section{\recompile implementation on top of FlexFlow}
\label{sec:rec-impl}
\begin{figure}[h]
    \includegraphics[width=0.485\textwidth]{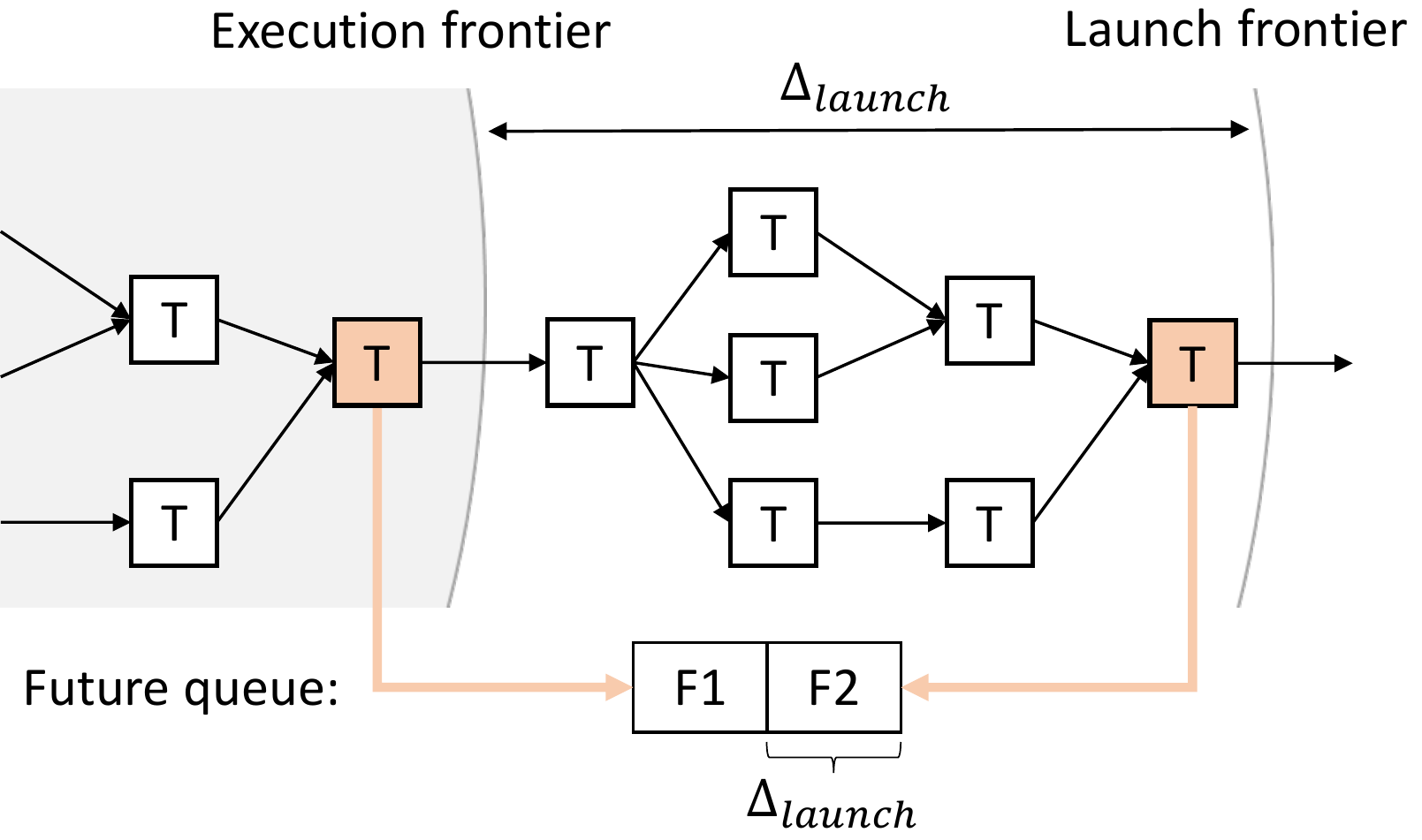}
    \caption{$\Delta_{launch}$ indicates the number of forward/backward iterations between the launch and execution frontiers and is a hyperparameter set by users (here, $\Delta_{launch}=1$). Like this, users can precisely control how many iterations after a certain event the recompilation should take effect.}
    \label{fig:frontiers}
\end{figure}

FlexFlow \cite{flexflow} is a distributed DNN framework built on top of the Legion programming system \cite{legion}.
One of the aspects of the Legion design is that it is highly asynchronous, launching tasks potentially well ahead of when they can be actually executed.   This gives rise to a \textit{launch frontier} along which tasks are launched, and an \textit{execution frontier} along which tasks are executed. Figure \ref{fig:frontiers} visualizes these two frontiers -- note that tasks between the two frontiers have been launched but haven't yet been executed.

When changing the computation graph of a model, the changes will only be reflected in tasks that haven't been launched yet --- all launched tasks still use the previous original computation graph. As a result, adjustments to a computation graph take effect at the launch frontier. 

However, graph adjustments are triggered based on the model metrics computed at the execution frontier. This creates a disparity, where graph adjustments only take effect with some delay after they are triggered. 

\recompile leverages this disparity as a time buffer during which the execution of the user's \recompile trigger and model training are overlapped.
During the execution of the \recompile trigger, no new tasks can be launched since it is unknown what tasks to launch. Without the buffered tasks that reside between the two frontiers (are launched but not executed), the model's training/inference would thus also need to be stalled until new tasks can be launched (i.e. the \recompile trigger has finished execution). However, as long as some launched tasks remain to be executed, stalling task launching does not stall task execution.

Therefore, it is crucial to launch enough tasks before executing a \recompile trigger, such that the execution of the trigger function finishes before all launched tasks have been executed. However, launching too many tasks in advance can result in the adjustment only taking effect after a long delay, which could lead to less speed up or decreased model quality. It is therefore critical that \recompile permits users to  specify how many tasks to buffer before executing a \recompile trigger.


In our \recompile implementation, users specify the number of tasks to buffer as a number of training/inference iterations $\Delta_{launch}$. \recompile  ensures that $\Delta_{launch}$ iterations worth of tasks are launched before a \recompile trigger is executed.
To enforce that exactly $\Delta_{launch}$ training/inference iterations lie between the execution and launch frontiers, \recompile maintains a queue for each model metric.  When launching a task that returns a model metric, a future for that metric is pushed to the queue. This future is a placeholder for the metric's value, which will only be known once the task has been executed. Once the task has been executed, \recompile will pop the corresponding future from the queue. Since a model metric is computed exactly once per iteration, the queue's length exactly indicates how many training/inference iterations have been launched but not executed.

\recompile can now simply enforce that the length of all Future queues corresponds exactly to $\Delta_{launch}$ before executing the \recompile trigger. If the queue is too long, \recompile waits until the right amount of model metrics have been computed (finished execution) and then calls the \recompile trigger with the most recently computed model metrics. If the queue is too short, \recompile simply launches more tasks before executing the trigger. The queue is only too short in two scenarios: In the beginning and end of training, and when the user increases $\Delta_{launch}$. Note that $\Delta_{launch}$ can always be changed at any point during runtime.



In Figure \ref{fig:frontiers}, $\Delta_{launch}$ is 1 iteration and \dynamoe waits on Future \verb|F1| before it will continue to launch tasks and eventually add Future \verb|F3|. In Figure \ref{fig:frontiers}, the seven tasks between the execution and the launch frontier correspond to one iteration, which is indicated by the top arrow labeled with $\Delta_{launch}$. In the moment depicted in Figure \ref{fig:frontiers}, the task returning the value of \verb|F1| has just been executed, which means that \verb|F1| will be popped from the queue and further tasks will be launched. Waiting on \verb|F1| made sure that \dynamoe doesn't launch more tasks than $\Delta_{launch} = 1$ iterations ahead.

\section{Support for \recompile in Other Frameworks}
\label{sec:rec-other}

\recompile can also be implemented on top of other existing ML frameworks. This section shows that these frameworks currently do not fulfill the requirements discussed in Section \ref{sec:recompile-overview} and would thus also benefit from supporting a mechanism like \recompileNoS. For reference, the requirements are again stated in the following:

\begin{enumerate}
    \item Users can monitor \textit{model metrics} with negligible overhead. A model metric can be anything that informs the user about the current state of the model during inference or training, such as training loss.
    \item Users can alter the computation graph with little overhead. When and how the graph is altered is determined by a user-provided \textit{recompile trigger}, which may base its decision on model metrics.
    \item Executing the \recompile trigger and computing the hardware mapping of the new graph is overlapped with inference/training using the current graph. For example: A user can conduct an expensive computation to decide how to alter the graph or conduct a long search for a new parallelization strategy while model training continues in parallel.
\end{enumerate}

Since \recompile only requires the ability to perform \recompile computations on CPUs in parallel with the training of an MoE model on GPUs, \recompile can be implemented for both static and dynamic computation graphs. \dynamoe chooses to implement \recompile for a static computation graph, which lets it combine the best of both worlds: Static computation graphs are generally faster since they allow for runtime optimizations that are impossible for dynamic graphs. We show that \dynamoe outperforms an MoE system that uses PyTorch's dynamic computation graph. Nevertheless, a static computation graph with \recompile still allows for dynamic graph adjustments --- we will later show that these adjustments even go beyond what is possible in frameworks like PyTorch.

In the following, we will consider to what extent TensorFlow\footnote{Without eager execution} and PyTorch are able to fulfill the three requirements in Section~\ref{sec:recompile-overview}. While other popular frameworks are not discussed in detail, PyTorch and TensorFlow should serve as general examples of the most popular frameworks using a static and a dynamic computation graph respecitively. Other popular frameworks also fall into one of these two categories: Caffe, Jax, Theano, and CNTK employ static computation graphs while Chainer and DyNet employ dynamic computation graphs \cite{caffe, jax, theano, cntk, chainer, dynet}.

\paragraph{\recompile in PyTorch.}
While PyTorch is able to fulfill the first two requirements, it fails to fulfill the last one.
\begin{enumerate}
    \item Model metrics can directly be monitored in PyTorch due to its support for dynamic model construction. PyTorch's API is hereby even more intuitive than the one of \dynamoe: while \recompile triggers and graph adjustments are defined as functions in \dynamoe, users can simply interleave them with the graph construction code in PyTorch. 
    \item PyTorch supports dynamic modifications of graph components during runtime. 
    \item PyTorch allows to interleave the graph execution with execution of code that triggers or performs graph adjustments (\recompile trigger). Since PyTorch launches GPU operators asynchronously, PyTorch can also execute \recompile triggers in parallel to model training/inference on GPU. However, PyTorch does not expose an API that allows \recompile triggers to use model metrics that are computed $\Delta_{launch}$ training/inference iterations before, where $\Delta_{launch}$ can be adjusted dynamically. Instead, using model metrics in the \recompile trigger would insert a dependency to a value computed in the current training/inference iterations, which would leave little time for the \recompile trigger to execute before GPU execution stalls. In order to support \recompile optimizations, a similar mechanism as in section \ref{sec:rec-impl} would therefore need to be implemented, for example using \verb|Torch.Futures|.

    
    \end{enumerate}

\paragraph{\recompile in TensorFlow.}
TensorFlow only fulfills the first requirement but fails to fulfill the latter two.

\begin{enumerate}
    \item TensorFlow provides {\em callbacks} for monitoring model metrics~\cite{tf-callback} in a similar way as \dynamoe. 
    However, TensorFlow does not support graph adjustments in a callback function, resulting in inability to support the second requirement.
    \item Graph adjustments in TensorFlow can only occur outside of a session. In order to perform a recompilation during training, users need to split their training procedure into multiple runs of individual Sessions between which a recompilation can be issued. This is infeasible, since it would require checkpointing and reloading the entire model every time a new session is run --- supporting recompilations at the temporal granularity of \dynamoe, would require storing and loading the entire model after every training batch iteration.
    \item Executing custom user code during graph execution also requires splitting training into many small sessions with user code executed between sessions. There is no ability to overlap computation; graph execution comes to a halt while user code runs, and the overhead from using many small sessions would be an additional performance penalty.
\end{enumerate}

In conclusion, TensorFlow and PyTorch do not natively fulfill the three requirements imposed on \recompile. They can thus not support optimizations like the ones proposed in Section~\ref{sec:recompile}, and would benefit from also implementing \recompile.

\section{Effect of caching on accuracy}
\label{sec:cache_acc}
\begin{figure}[h]
    \centering
    \includegraphics[width=0.43\textwidth]{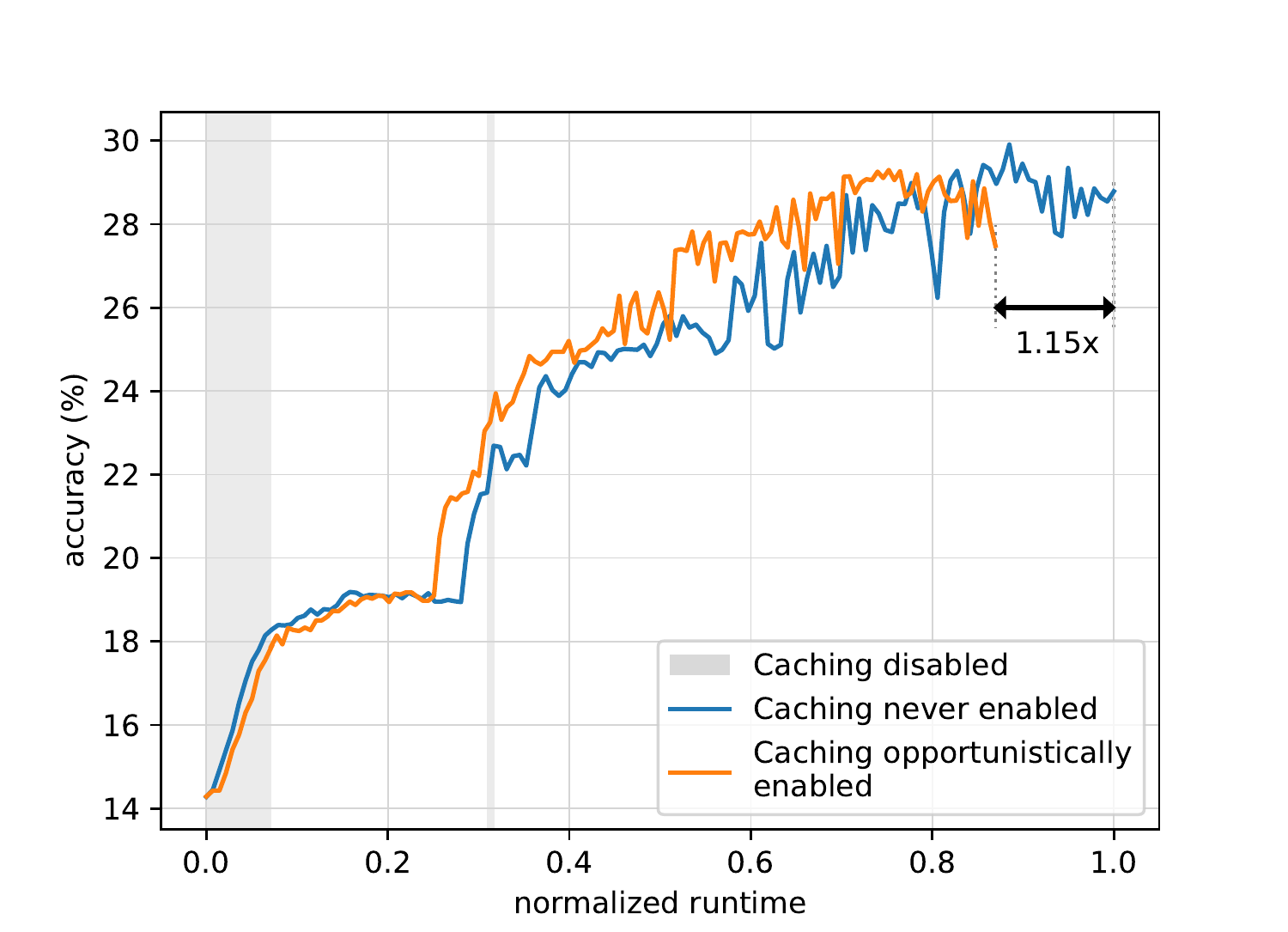}
    \caption{Enabling caching has little effect on the training progress and yields a $1.15\times$ speed up.}
    \label{fig:cacheacc}
\end{figure}

Figure \ref{fig:cacheacc} shows the accuracy progression when adaptively enabling caching over large parts of the training or disabling it over all of the training. Note that the x-axis depicts runtime and not traiing epochs. Both models were run for the same amount of training epochs.

\end{document}